\documentclass{ieeetj}
\usepackage{amsmath,amssymb,amsfonts}
\usepackage{graphicx,color}
\usepackage{textcomp}
\usepackage{hyperref}
\hypersetup{hidelinks}
\usepackage{algorithm,algorithmic}
\def\BibTeX{{\rm B\kern-.05em{\sc i\kern-.025em b}\kern-.08em
    T\kern-.1667em\lower.7ex\hbox{E}\kern-.125emX}}
\AtBeginDocument{\definecolor{tmlcncolor}{cmyk}{0.93,0.59,0.15,0.02}\definecolor{NavyBlue}{RGB}{0,86,125}}

\renewcommand{\appendixname}{APPENDIX}
\usepackage{bm}
\usepackage{booktabs}
\usepackage{multirow}
\usepackage[most]{tcolorbox}
\usepackage{xcolor} 
\def\OJlogo{\vspace{-4pt}}
\def\seclogo{\vspace{10pt}$<$Society logo(s) and publication title will appear here.$>$}
\def\authorrefmark#1{\ensuremath{^{\textbf{#1}}}}
\begin{document}

% --- arXiv safe: avoid math mode in header logo ---
\makeatletter
\renewcommand{\OJlogo}{\vspace{-4pt}} % no $ or $$, no math mode
\makeatother

%\receiveddate{XX Month, XXXX}
\reviseddate{May, 2025}
%\accepteddate{XX Month, XXXX}
%\publisheddate{XX Month, XXXX}
%\currentdate{XX Month, XXXX}
%\doiinfo{XXXX.2022.1234567}

\markboth{}{Author {et al.}}

\title{Generative Agents for Innovation}
\author{Masahiro Sato\authorrefmark{1}}
\affil{Tohoku University, Sendai, Miyagi 980-8576, Japan}
\corresp{Corresponding author: Masahiro Sato (email: masahiro.sato@tohoku.ac.jp).}
\authornote{This work was supported by JSPS KAKENHI (Grant Numbers: 21K13276 and 25K05330), CAINZ Digital Transformation Foundation, and Telecommunication Advancement Foundation.}

\begin{abstract}
This study examines whether collective reasoning among generative agents can facilitate novel and coherent thinking that leads to innovation. To achieve this, it proposes GAI (Generative Agents for Innovation), a new LLM-empowered framework designed for reflection and interaction among multiple generative agents to replicate the process of innovation. The core of the GAI framework lies in an architecture that dynamically processes the internal states of agents and a dialogue scheme specifically tailored to facilitate analogy-driven innovation. The framework’s functionality is evaluated using Dyson's invention of the bladeless fan as a case study, assessing the extent to which the core ideas of the innovation can be replicated through agents' dialogue using a set of fictional technical documents. The experimental results demonstrate that models with internal states significantly outperformed those without, achieving higher average scores and lower variance. Notably, the model with five heterogeneous agents equipped with internal states successfully replicated the key ideas underlying the Dyson's invention. This indicates that the framework enables agents to refine their ideas, resulting in the construction and sharing of more coherent and comprehensive concepts.
\end{abstract}

\begin{IEEEkeywords}
Generative AI, Innovation, Large language models, Multi-Agent Systems, Patents
\end{IEEEkeywords}


\maketitle
\section{INTRODUCTION}
Generative agents, including LLM agents, can perform various complex tasks by breaking them down, conducting systematic reasoning, leveraging external resources, learning from experiences, and planning for future. LLM-based Multi-agent System (LLM-MAS) extends this capability by incorporating collective reasoning functions, with applications already emerging in fields such as software development, urban planning, scientific debates, and social simulations (\cite{ataei2024elicitron, chan2023chateval, chen2025multiagentconsensus, chen2024agentverse, du2023, chuang2024opinion, gao2024, hong2024metagpt, huang2025howfar, kaiya2023LyfeAgents, CAMEL, Li_2023ToM, Li2024LongShortTermReasoning, li2023metaagents, liang2024encouraging, lin2023agentsims, liu2024DyLAN, Omirgaliyev, park2023generativeagents, qian2024chatdev, Shinn2023, slumbers2024leveraging, tang2024medagents, wang2024unleashing, williams2023epidemic, wu2023autogen, Xiong_2023, yang2024xagents, Zhang2024Recommendation, zhang2024simple, zhang2024cumulativereasoning, zhou2024sotopia}). Applying these capabilities of generative agents to discussions based on technical documents, such as patents, could potentially replicate the collective interactions that human teams engage in during innovation processes.

Efforts to analyze innovation processes have historically included observational experiments in actual corporate environments, but these have faced significant limitations due to low reproducibility and difficulties in manipulating experimental conditions. As a result, the effects of conditions such as diversity in team member, communication methods, and organizational structure on innovation processes remain not fully understood. If LLM-MAS can replicate the innovation processes of human teams, it would enable repeated simulations under adjustable conditions with high reproducibility.

Furthermore, artificial intelligence technologies, including LLMs, will increasingly be applied in real-world corporate innovation settings (\cite{Barbieri02072024, Bilgram2023, Bouschery2023, Boussioux2023, cimino2024generative, CORVELLO2025100456, geyer2024influence, gindert2024, HAO2024102662, harwood2023, HOLMSTROM2024, MARIANI2024114542, ROBERTS2024103081, sedkaoui2024generative, SINGH2024103021, singh2024, URBAN2024105031, WAELALKHATIB2023102403, yeverechyahu2024}). If LLM-MAS can simulate innovation processes, it would also be possible to pioneer this new domain by integrating generative agent teams with human teams, demonstrating new possibilities for AI-human collaboration in driving innovation.

The objective of this study is to examine whether collective reasoning among generative agents can facilitate novel and coherent thinking that leads to innovation. To achieve this, the study focuses on \textit{analogy-driven innovation} and proposes GAI (Generative Agents for Innovation), a new LLM-empowered framework designed for reflection and interaction among generative agents to replicate this process.

The core of the GAI framework lies in an agent's architecture that dynamically processes its internal state, and a dialogue scheme specifically designed to facilitate analogy-driven innovation. In the former, each agent generates and refines ideas based on intrinsic motivation and critical introspection, leveraging previously reviewed documents and interactions with others. This process forms and updates the current thought of the agent, allowing the agent’s contributions to the dialogue to dynamically self-evolve in a coherent way over time. In the latter, the scheme incorporates insights from management and cognitive science to replicate the process of identifying functional similarities and mechanical differences, enabling the transfer of solutions across various technological domains.

To evaluate the functionality of the framework, this study performs two experiments using a set of fictional technical documents. The one is to take Dyson's invention of the bladeless fan as a case study and examine the extent to which the innovation can be replicated using household fans as the target domain and industrial ejectors as the source domain. The other is to apply the framework to ten different source domains to examine its applicability.

\section{PROPOSED FRAMEWORK}

\subsection{AGENT ARCHITECTURE}
\begin{figure*}
    \centering
    \includegraphics[width=\textwidth]{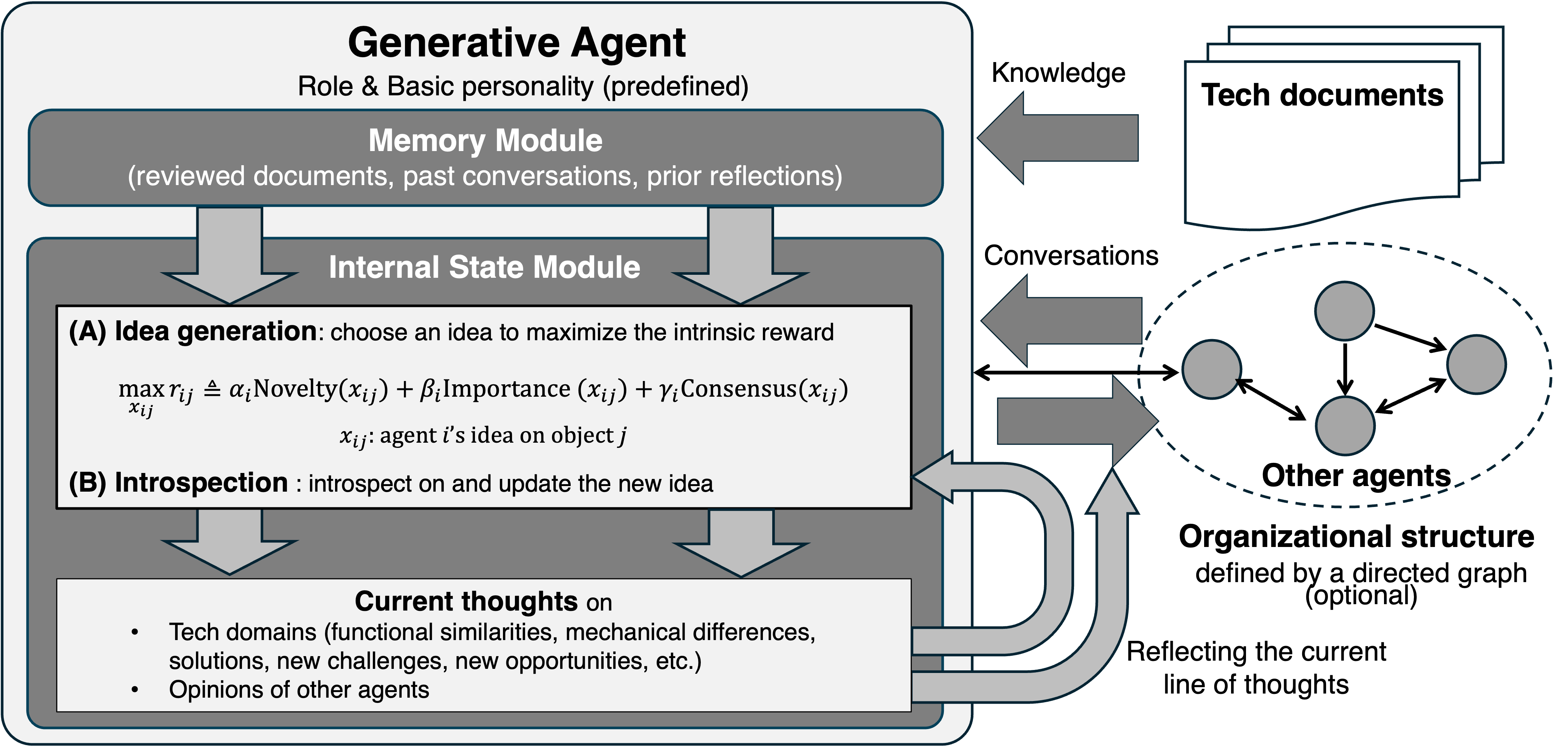}
    \caption{Agent architecture and organizational structure}
    \label{fig:arch}
\end{figure*}

The architecture of an agent consists of two primary LLM-driven modules: a memory module and an internal state module (Figure \ref{fig:arch}). The \textit{memory module} records a) contents of reviewed technical documents, b) past conversations with others, and c) the agent's prior reflections, all timestamped for reference. These records are then organized into structured memory, which is input into the internal state module. 

The \textit{internal state module} processes newly acquired technical information and statements from others during the dialogue and facilitates innovative and coherent thinking of an agent. Specifically, taking in external inputs from technical documents and other agents, as well as internal inputs from the agent's prior reflections and its latest thoughts, the module continuously updates the agent’s \textit{current thoughts}, which is a collection of texts structured according to the dialogue scheme for analogy-driven innovation. Current thoughts recurrently serve as the foundation for future reflections of the agent and function as the basis for the agent's statement in a dialogue, providing it with strong context awareness and coherency.

When forming and updating current thoughts, the internal state module undergoes two processes: idea generation and introspection. During the \textit{idea generation} process, the module is prompted to generate multiple ideas distinct with each other for reflections on a technological domain based on the memories retrieved from the memory module and the agent's current thoughts at that time. Subsequently, the module initiates another LLM instance and prompts it to evaluate these ideas by an integer score on a scale of 1 to 10 for each of three criteria:
\begin{itemize}
    \item \textbf{Novelty}: the degree to which the generated idea introduces new content that has not been present in prior conversations and reflections
    \item \textbf{Importance}: the degree to which the generated idea is significant in light of previous discussions and reflections
    \item \textbf{Consensus}: the degree to which the generated idea aligns with the opinions of the majority of participants, based on past conversations
\end{itemize}

The module selects the idea that maximizes an intrinsic reward calculated as a weighted sum of these criteria: 
\begin{equation}
\begin{split}
    \max_{x_{ij}} r_{ij} &\equiv \alpha_i \, \text{Novelty}(x_{ij}) + \beta_i \, \text{Importance}(x_{ij}) \\
    & + \gamma_i \, \text{Consensus}(x_{ij}), \label{eq:max_reward}
\end{split}
\end{equation}

\noindent
where $r_{ij}$ represents the intrinsic reward of agent $i$, $x_{ij}$ is its idea on object $j$, and $\alpha_i$, $\beta_i$, and $\gamma_i$ respectively represent the weights for the 1-10 integer scores on each criterion, which are denoted by Novelty$(x_{ij})$, Importance$(x_{ij})$, and Consensus$(x_{ij})$ in Equation \eqref{eq:max_reward}. By assigning different weights to these criteria for different agents, the framework can represent diverse intrinsic motivations. For instance, an agent with a higher weight on Novelty will prioritize curiosity and innovation, whereas an agent with a higher weight on Consensus will emphasize collaboration and agreement.

During the \textit{introspection} process, the selected idea is critically examined to identify any technical ambiguities or contradictions. Specifically, the module sets up another LLM instance as an \textit{internal critic} and prompts it to examine if there are any important points in the selected idea that are unclear from the perspective of technology, and if there are any contradictions from the perspective of technology. The internal state module is then prompted to update the idea by including responses that accurately address the questions and points of contradiction raised through the introspection, refining the idea to improve clarity and coherence.

The internal state module not only reflects on the agent's own thoughts about technological domains but also evaluates the opinions expressed by other participants. Specifically, based on prior conversations and the agent’s current thoughts, the module is prompted to identify aspects of each participant’s opinions that the agent agrees with or finds interesting, as well as those that it considers contradictory or lacking. These reflections on others are then reflected in the agent's current thoughts together with its own ideas. 

Through these steps, the agent's current thoughts are continuously updated, incorporating refined ideas and reflections on domains and others. During the dialogue, the agent makes statements based on its current thoughts, shaped through these processes, as well as on memories retrieved from the memory module. Even if agents are initially assigned homogeneous roles and personality, their thoughts dynamically evolve in distinct ways through these processes.

\subsection{DIALOGUE SCHEME FOR ANALOGY-DRIVEN INNOVATION}

Innovation takes various forms, such as technology-driven innovation, market-driven innovation, open or closed innovation. This paper focuses on \textit{Design by Analogy} (DbA), an approach to innovation that has been extensively studied across disciplines, including management science and cognitive science (\cite{fu2013meaning, fu2015design, gassmann2008opening, goel1997design, holyoak1994, hope2017accelerating, huhns1988argo, jeong2014creating, linsey2008increasing, linsey2008modality,  mcadams2002quantitative, Melnychuk2020, moreno2014fundamental, murphy2014function, song2019design, song2022design}). Many groundbreaking innovations, such as the application of racing car technologies to running shoes and various forms of biomimicry, employ cross-domain analogies. DbA addresses problems in a target domain by extracting and transferring design solutions from other domains (source domains). The central focus of this approach lies in identifying similarities and differences. In particular, it is widely recognized that superior innovations abstract away from surface-level or attribute-based similarities, instead relying on functional and relational similarities for analogy (\cite{bhatta1996design, fu2015design, fu2013meaning, holyoak1994, linsey2008modality, mcadams2002quantitative}).

Building on these interdisciplinary insights, this study proposes a dialogue scheme that leverages collective reasoning among agents to replicate the process of DbA, extracting functional similarities and mechanical differences from technical documents and transferring solutions across domains.

\begin{figure*}
    \centering
    \includegraphics[width=\textwidth]{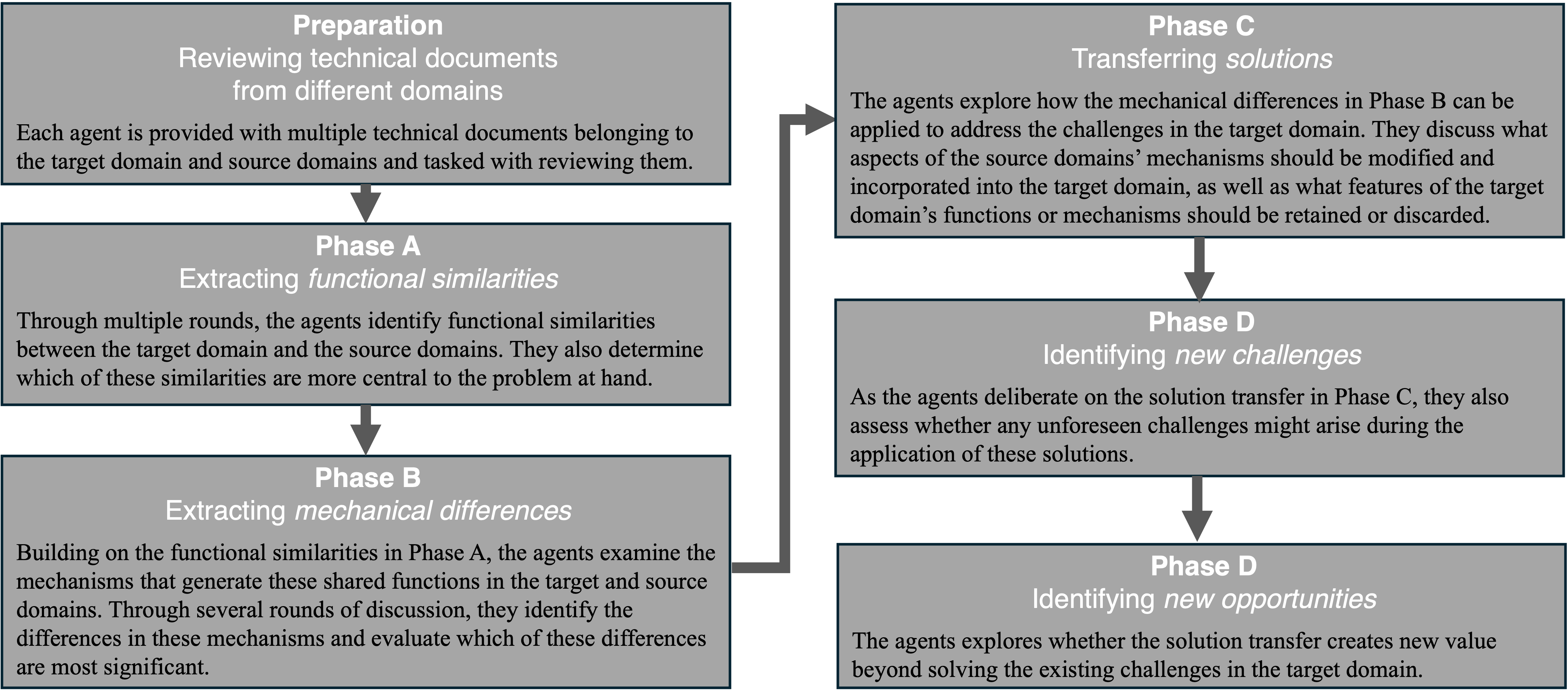}
    \caption{Dialogue scheme for analogy-driven innovation}
    \label{fig:dialogue}
\end{figure*}

Specifically, this scheme facilitates the procedure shown in Figure \ref{fig:dialogue}.

\subsection{ORGANIZATIONAL STRUCTURE}
The organizational structure of the agent team is defined as a directed graph, with agents as vertices and the directions of communications or instructions as edges. Each agent can only communicate with adjacent agents and issue instructions in the direction of the edges. By utilizing this directed graph structure, the framework can represent various organizational forms, ranging from flat to hierarchical structures, and evaluate their impact on technological discovery.

\section{EXPERIMENTS}

\subsection{CASE: Dyson Air Multiplier}

In household fans, it is common to cover the spinning blades with a grid-like guard to prevent accidental contact. However, such guards not only reduce the efficiency and consistency of airflow but also make the device less convenient to carry and clean. Dyson Air Multiplier addressed these challenges applying the principles of fluid dynamics used in jet engines and achieving a transition from a physical-blade-based airflow mechanism to a fluid-dynamics-based airflow mechanism leveraging pressure differential.

Air Multiplier is often referred to as a "bladeless fan," but this is somewhat misleading (\cite{hallgeisler2022}). In reality, it uses blades enclosed within the base, powered by an efficient brushless motor to draw in air. Atop the base, a ring-shaped airfoil with a diameter of approximately 30 cm is mounted. Air drawn into the device is expelled through narrow slits of the jet nozzle on the inner surface of the ring. During this process, the fan pulls in air from behind the ring, effectively "multiplying" the airflow (\cite{mansson2014flow}). Specifically, the air drawn from the base is accelerated through the jet nozzle, creating a pressure gradient that induces surrounding air to be pulled in (inducement). Additionally, the high-speed expelled air entrains the surrounding air, generating a consistent airflow directed forward from the ring (entrainment).

Dyson introduced the first generation of Air Multiplier to the market in 2009. However, it faced challenges with noise levels. To address this issue, Dyson invested over \$60 million in further R\&D, culminating in the release of the second generation in 2014,  which incorporated Helmholtz cavities into the base, effectively resolving the turbulence issues that were a primary source of noise (\cite{guardian2014, hallgeisler2022}).

\subsection{EXPERIMENT 1}
In the experiment, dialogues are conducted with household fans as the target domain and industrial ejectors as the source domain. Ejectors are industrial devices used for transporting fluids (liquids or gases). They are utilized not only for water supply, exhaust, and smoke extraction but also for neutralization, dilution, and cooling through the mixing of gases or liquids, as well as for creating vacuum conditions as a vacuum pump.

Although the applications are entirely different, the tow domains share a functional similarity, the movement of fluids. However, the mechanisms that enable this function are entirely distinct. In a typical ejector, a nozzle draws high-speed driving fluid into the body as a driving source. As the high-speed driving fluid reduces the pressure inside the body, creating negative pressure, the surrounding low-speed external fluid is pulled into the body. The mixed fluid of the driving and external fluids is then discharged outside the body through a diffuser. This allows the ejector to move external fluid without any external power sources.

In the practical stage of the proposed framework, actual technical documents, such as patent filings, are expected to be used. However, in the experimental environment presented in this paper, fictional technical documents based on real-world technical information are prepared, primarily considering budgetary constraints. These fictional documents, like actual patent documents, include miscellaneous information that does not necessarily lead to the target invention. For instance, the document for ejectors contains information about a wide range of industrial applications far removed from household use, details about unrelated components, and data on various types of irrelevant fluids and materials. Similarly, the document for household fans include various unrelated information. Agents must extract fragments of information that lead to a "solution" from this broad and miscellaneous data and connect those fragments using their independent reasoning. In this particular experiment, to reach the target innovation, agents are required to reinterpret the concept of driving fluid from the perspective of a household fan and devise a method to generate it within the constraints of household technology and energy environments. However, the agents are prohibited, by prompts, from using pre-trained knowledge about Dyson products or any information available on the internet. Instead, they are required to base their statements and reflections strictly on logical conclusions or associations derived from their prior discussions and reflections. 

The evaluation is conducted for the solutions derived from the dialogue using a point-based system based on the nine criteria listed in Table \ref{table:criteria} (maximum of 8 points). These criteria are developed from the perspective of whether the transfer of solutions comprehensively and coherently covers the technological elements corresponding to the first-generation Dyson Air Multiplier. Since the second generation requires computational fluid dynamics simulations for the specific shape of the fan to address the turbulence issues, they were excluded from the evaluation in this experiment, which only utilizes language models.

\begin{table}
    \centering
    \caption{Evaluation criteria}
    \begin{tabular}{cp{7.4cm}}
        \toprule
        No. & \multicolumn{1}{c}{Description} \\ \hline
        1 & If no direct mention of blades is made, is a realistic and safe device capable of generating driving fluid within the size and price range of household appliances proposed? (2 points) \\ \hline
        2 & Is retaining blades proposed, and are they identified as playing a role in generating driving fluid or negative pressure? (1 point) \\ \hline
        3 & Is retaining blades proposed, and is enclosing them inside the main body suggested? (1 point) \\ \hline
        4 & If no direct mention of a motor is made, is retaining blades proposed, or is the device proposed in criterion 1 conventionally paired with a motor? (1 point) Additionally, is the use of household electricity as the motor’s power source mentioned? (1 point) \\ \hline
        5 & Is retaining a motor proposed, and is it identified as playing a role in generating driving fluid or negative pressure? (1 point) \\ \hline
        6 & Are advanced proposals made regarding a motor to improve its efficiency or noise reduction, such as the use of a brushless motor? (1 point) \\ \hline
        7 & Is the introduction of a nozzle structure proposed? (1 point) \\ \hline
        8 & Other than the nozzle, are any proposals made to optimize internal channels for the efficiency or noise reduction of the driving fluid? (Up to 2 points depending on the number and specificity of the proposals) \\ \hline
        9 & Are relevant principles of fluid dynamics explicitly explained? (1 point) \\
        \bottomrule
    \end{tabular}
    \label{table:criteria}
\end{table}

To evaluate the performance of the proposed framework, the results are compared across eight models shown in Table \ref{table:8cases}, differentiated based on three perspectives: (1) the number of agents, (2) the presence of internal states, and (3) the heterogeneity of intrinsic motivations. The first one examines the impact of increasing the number of agents. When multiple agents are involved, the evaluation is conducted based on the majority opinions. The second one assesses the functionality of the internal state module. In models without internal states, agents are required to make statements based solely on their initially assigned roles and personalities, as well as the history of past conversations, without undergoing the process of reflection. The third one evaluates the effect of heterogeneity among agents. In homogeneous models (Models D and G), all agents equally consider the three intrinsic motivations (novelty, importance, and consensus). In heterogeneous models (Models E and H), agents are configured with different emphases: some prioritize novelty while de-emphasizing consensus, others prioritize consensus while de-emphasizing novelty, and some consider all three motivations equally. The experiment is conducted 10 times for each model using a common set of 10 seeds. The ratings are done by carefully examining the generated solutions through human assessment.

\begin{table}
    \centering
    \caption{Eight models for experiment}
    \begin{tabular}{lccc}
        \toprule
        \multicolumn{1}{c}{Model} & \# of agents & Internal state & intrinsic motivation \\ \hline
        Model A & 1 & No & n.a.\\
        Model B & 1 & Yes & n.a.\\ \hline
        Model C & 3 & No & n.a.\\
        Model D & 3 & Yes & homogeneous\\
        Model E & 3 & Yes & heterogeneous\\ \hline
        Model F & 5 & No & n.a.\\
        Model G & 5 & Yes & homogeneous\\
        Model H & 5 & Yes & heterogeneous\\
        \bottomrule
    \end{tabular}    
    \label{table:8cases}
\end{table}

\subsection{EXPERIMENT 2}
To examine the applicability of the framework, the second experiment applies it to ten different source domains: (1) safety sensors, (2) cooling mats, (3) shock-absorbing materials, (4) air coolers, (5) aerogenerators, (6) mist spray cooling systems, (7) small hydropower systems, (8) noise-canceling technology, (9) airbags, and (10) vacuum cleaners. These domains are selected because they share certain functions with household fans, such as cooling effects and energy utilization through rotational motion, or because they relate to key issues in household fans, such as injury prevention, noise reduction, and energy efficiency. The list of the ten domains and the reasons for selection is given in Table \ref{table:10domains}. The experiment is conducted for two extreme models: the sigle-agent model without an internal state (Model A) and the five heterogeneous agent model with internal states (Model H).

\begin{table}
    \centering
    \caption{Ten source domains}
    \begin{tabular}{p{2.2cm}p{5cm}}
        \toprule
        Domain & Reason for selection \\ \hline
        Safety sensors & Related to key issues in household fans (injury prevention) \\ \hline
        Cooling mats & Share functions with household fans (cooling effect) \\ \hline
        Shock-absorbing materials & Related to key issues in household fans (injury prevention) \\ \hline
        Air coolers & Share functions with household fans (cooling effect) \\ \hline
        Aerogenerators & Share functions with household fans (energy utilization through rotational dynamics) \\ \hline
        Mist spray cooling systems & Share functions with household fans (cooling effect) \\ \hline
        Small hydropower systems & Share functions with household fans (energy utilization through rotational dynamics) \\ \hline
        Noise-canceling technology & Related to key issues in household fans (noise reduction) \\ \hline
        Airbags & Related to key issues in household fans (injury prevention) \\ \hline
        Vacuum cleaners & Share functions with household fans (energy utilization through rotational dynamics) \\
        \bottomrule
    \end{tabular}    
    \label{table:10domains}
\end{table}

\vspace{1em}
In both experiments, the language model used to power the framework is OpenAI's GPT-4o. To ensure uniform experimental conditions, word limits are imposed on the agents' statements and reflections in each round. In this preliminary experiment, a flat organizational structure was adopted, where all agents could communicate with all others. Other structures, such as hierarchical models, are left for future research.

\subsection{RESULTS}

Table \ref{table:results_1} summarizes the results of Experiment 1. In all cases with 1, 3, and 5 agents, the models with internal states outperformed those without on average. This indicates that, through refinement by internal states, more coherent and comprehensive ideas are constructed and shared among agents. For example, in the single-agent model without an internal state (Model A), the proposal of retaining blades or incorporating an appropriate blade-driven device as a mechanism to generate a driving fluid (Criteria 1 and 2 in Table \ref{table:criteria}) was made only 3 times out of 10. In contrast, in the single-agent model with an internal state (Model B), such mechanisms were proposed 9 times out of 10. This indicates, in the latter model, the concept of external driving fluid was successfully reframed within the context of household fans as the intake of surrounding air by blades. Furthermore, in the model without an internal state, aside from the introduction of nozzle structures, the proposal for optimizing internal channels to improve the efficiency and reduce the noise of the driving fluid (Criterion 8) was made only once out of 10 times. In contrast, in the model with an internal state, such proposals were made 7 times out of 10, ranging from the redesign of the fan housing to the introduction of guide vanes and the use of Computational Fluid Dynamics (CFD) simulations. Below is an example of part of an agent's statement in the single-agent model with an internal state (Model B).

\vspace{1em}
\noindent
\begin{tcolorbox}[
    colback=gray!20,
    colframe=black,
    boxrule=0.2mm,
    breakable,
    enhanced,
    sharp corners
]
\small \textbf{Agent\_1:} Rather than relying solely on large rotating blades, we might consider smaller impellers driven by the existing motor, which would introduce directed air streams. ... Retaining a compact motor system that uses lightweight, efficient turbines is crucial, as it allows for consistent airflow without excessive bulk. ... By strategically designing air inlets and channels, we can naturally guide air currents, even without a high-velocity driving fluid. Prioritizing safety, all moving parts should be enclosed within the casing, utilizing protective materials to eliminate the risk of contact with blades. ... Utilize computational fluid dynamics (CFD) to refine blade shapes and angles, enhancing air movement and improving efficiency. ... Incorporate guide vanes that naturally direct airflow, minimizing reliance on high-velocity inputs. This approach ensures smoother air pressure control and efficient cooling without adding mechanical complexity.
\end{tcolorbox}

In contrast, the impact of the number of agents is not straightforward. On one hand, the effect is generally positive for models without internal states. First, the average performance improved significantly when the number of agents increased from one (Model A) to multiple (Models C and F). Second, when comparing the three-agent model (Model C) to the five-agent model (Model F), increasing the number of agents significantly reduced the variance in scores, although the average scores remained unchanged. The larger variance in scores with three agents may partly stem from the stricter majority requirement for consensus, where the opinion of a single agent can heavily influence the results.

On the other hand, when the agents have an internal state, the single-agent model (Model B) outperformed almost all multi-agent models both with and without internal states (Models C, D, E, F, and G) except for Model H. It also recorded the smallest variance across all models. This highlights the strong capability of the internal state module in deriving comprehensive and coherent ideas, even when acting alone. However, its superior performance compared to some multi-agent models with internal states may also be attributed to the stricter majority requirement. The larger variance observed in the three-agent models suggests this.

The highest average score among all models was recorded by the case with five heterogeneous agents with internal states (Model H). Increasing the number of agents from 3 to 5 slightly relaxed the consensus requirement, allowing a broader range of opinions to emerge from more diverse agents, which were then refined and shared through collective reasoning. The positive effects of the number and diversity of agents are consistent with the results of some prior studies on LLM-MAAS (\cite{chan2023chateval, chen2025multiagentconsensus, du2023}). Below is an example of part of agents' statements in the five heterogeneous agent model with internal states (Model H).

\vspace{1em}
\noindent
\begin{tcolorbox}[
    colback=gray!20,
    colframe=black,
    boxrule=0.2mm,
    breakable,
    enhanced,
    sharp corners
]
\small \textbf{Agent\_1:} ... First, retaining the electric motor is crucial, but shifting the focus from blades to a ducted design with specialized nozzles could minimize safety risks and maintenance. ... This means adopting a circular or linear duct system to channel airflow, creating efficient fluid movement pathways. The key is to enhance the fan's ability to draw in and expel air smoothly using controlled high-speed airflow, perhaps with adjustable nozzles or diffusers that modify air velocity and direction. ...

\noindent
\small \textbf{Agent\_2:} ... However, traditional metal or plastic grids could be phased out if the airflow generation can be enclosed entirely within the fan's structure.

\noindent
\small \textbf{Agent\_3:} ... The most critical adaptation would be incorporating the ejector's pressure-driven flow to create a bladeless fan. This could be achieved using a small compressor or blower to generate a high-pressure air source, channeling it through a nozzle to produce a coherent airflow. ...

\noindent
\small \textbf{Agent\_5:} ... Retaining the motor to power a compressor or blower generating a directed airflow stream aligns with fluid dynamics principles, enhancing efficiency. ...

\noindent
\small \textbf{Agent\_1:} ... Noise minimization can be achieved by including sound-dampening materials and CFD-aided design to manage high-speed airflow noise impacts. ...

\end{tcolorbox}

The agents' dialogue above proposed several ideas, including eliminating the existing grid guard and conducting airflow generation within an enclosed environment inside the fan, adopting a ducted design to improve airflow efficiency, utilizing a motor-driven blower to generate high-speed driving fluid, and performing CFD simulations to reduce noise. These proposals demonstrate that the key ideas underlying Dyson Air Multiplier have been successfully replicated.

\begin{table*}
    \centering
    \caption{Results of Experiment 1}
    \begin{tabular}{lcccc}
        \toprule
        \multicolumn{1}{c}{Model} & Min & Max & Average score & Standard deviation \\ \hline
        Model A: 1 agent without an internal state & 1 & 5 & 2.4 & 1.58 \\
        Model B: 1 agent with an internal state & 4 & 7 & 5.1 & 0.99 \\ \hline
        Model C: 3 homogeneous agents without internal states & 1 & 7 & 3.7 & 2.45 \\
        Model D: 3 homogeneous agents with internal states & 2 & 7 & 4.4 & 2.07 \\
        Model E: 3 heterogeneous agents with internal states & 2 & 7 & 4.0 & 2.05 \\ \hline
        Model F: 5 homogeneous agents without internal states & 1 & 6 & 3.7 & 1.83 \\
        Model G: 5 homogeneous agents with internal states & 2 & 6 & 4.7 & 1.16 \\
        Model H: 5 heterogeneous agents with internal states & 3 & 7 & 5.3 & 1.16 \\
        \bottomrule
    \end{tabular}
    \label{table:results_1}
\end{table*}

Table \ref{table:results_2} summarizes the results of Experiment 2. The most significant difference between the two models is the levels of technological coherence. In the single-agent model without an internal state (Model A), several proposals for some source domains are highly inconsistent or misaligned with the intended applications. For instance, in the cases of cooling mats and shock-absorbing materials, the model suggests that incorporating these materials into the guard can reduce the risk of injury. Similarly, in the airbag case, it proposes a retractable blade system to store the blades when not in use, arguing that this could reduce the risk of accidental contact and potentially eliminate the need for a protective guard. Furthermore, in the safety sensors case, both models propose the introduction of an automatic stopping mechanism. However, while the five-agent model with internal states (Model H) suggests retaining the guard as a safety backup, the model without an internal state (Model A) considers the possibility of removing the guard entirely. In addition to the technological coherence, another noteworthy aspect of the five-agent model with internal states (Model H) is the diversity of proposals across multiple source domains. These include hybrid manual-automatic systems and the adoption of lightweight, durable materials, demonstrating a broader range of innovative solutions. 

In sum, the second experiment demonstrates that the five heterogeneous agent model with internal states outperforms the single-agent model without an internal state both in the levels of technological coherence and diversity of proposals.

\begin{table*}
    \centering
    \scriptsize
    \caption{Results of Experiment 2}
    \begin{tabular}{p{2.4cm}p{4.4cm}p{4.4cm}p{4.4cm}}
        \toprule
        \multirow{2}{*}{Domain} & \multirow{2}{*}{Common proposals} & \multicolumn{2}{c}{Major differences} \\
           & & \multicolumn{1}{c}{Model A} & \multicolumn{1}{c}{Model H} \\ \hline
        Safety sensors & Adaptive control systems that allow for autonomous real-time adjustments by smart sensors. Automatic stopping by proximity sensors. & The conventional guard design can be discarded or modified. & Maintaining a minimal guard design as a safety backup. \\ \hline
        Cooling mats & Passive cooling materials within fan structures to enhance cooling efficiency and reduce noise. & The conventional guard design can be discarded or modified. & Maintaining a minimal guard design as a safety backup. \\ \hline
        Shock-absorbing materials & Shock-absorbing materials in fan components to reduce noise. & Adopting hock-absorbing materials for blade guards can mitigate the risk of injury. & Adopting passive materials to absorb vibrations can enhance the fan's energy efficiency. \\ \hline
        Air coolers & Incorporating evaporative cooling components. & A bladeless design could be explored. & Retaining core air movement features like rotating blades. A detachable, modular evaporative units with lightweight, eco-friendly materials. \\ \hline
        Aerogenerators & Sensors and automated systems to optimize airflow dynamically. Adjustable, aerodynamically efficient blade designs. & & \\ \hline
        Mist spray cooling systems & Hybrid systems integrating adjustable misting functions with traditional fan mechanisms. Modular and detachable misting components. & & Smart sensors to automatically adjust mist output. Moisture-resistant materials and waterproof seals. \\ \hline
        Small hydropower systems & Aadaptive control systems that allow for autonomous real-time adjustments by smart sensors. Advanced blade aerodynamics from hydropower turbines to design fan blades. & & Hybrid control mechanisms combining manual and automatic modes. \\ \hline
        Noise-canceling technology & Adaptive control systems that allow for autonomous real-time adjustments by smart sensors. & Modifying blade materials to incorporate sound-dampening qualities. & Hybrid control mechanisms combining manual and automatic modes.  A modular design approach. Leverage machine learning to predict and adapt fan’s settings for optimal comfort. Motors utilizing principles from noise-canceling technology to reduce operational noise. \\ \hline
        Airbags &  & A retractable blade system that allows blades to retract into a housing unit when not in use, minimizing the risk of accidental contact. Automatic stopping mechanism by proximity sensors. The bulky guards could potentially be discarded in favor of these advanced safety mechanisms. & Adaptive control systems that allow for autonomous real-time adjustments by smart sensors. Hybrid control mechanisms combining manual and automatic modes. Adopting lightweight, durable materials. \\ \hline
        Vacuum cleaners & Transitioning to enclosed or bladeless designs by reimagining the internal airflow pathway. Incorporating adjustable components that can customize airflow (e.g., directional louvers or nozzles). Adopting filtration technologies. & & Advanced brushless motor systems. \\
        \bottomrule
    \end{tabular}
    \label{table:results_2}
\end{table*}

\section{RELATED WORKS}
\subsection{INNOVATION and GENERATIVE AI}
Numerous studies have examined the impact of generative AI on corporate innovation, particularly in the field of management science (\cite{Barbieri02072024, Bilgram2023, Bouschery2023, Boussioux2023, cimino2024generative, CORVELLO2025100456, geyer2024influence, gindert2024, HAO2024102662, harwood2023, HOLMSTROM2024, MARIANI2024114542, ROBERTS2024103081, sedkaoui2024generative, singh2024, SINGH2024103021, URBAN2024105031, WAELALKHATIB2023102403, yeverechyahu2024}). These studies include analyses of the roles of generative AI in innovation management (\cite{CORVELLO2025100456, MARIANI2024114542, sedkaoui2024generative}), experiments using generative AI in human's innovation processes (\cite{Barbieri02072024, Bilgram2023, Boussioux2023, gindert2024, harwood2023, URBAN2024105031, yeverechyahu2024}), the impact on human decision-making (\cite{HAO2024102662}), the acceptance of generative AI (\cite{cimino2024generative}), and the classification of innovation strategies (\cite{HOLMSTROM2024}). Additionally, numerous studies have explored the use of generative AI in patent analysis and technology assessment, such as patent mapping and technology opportunity evaluation (\cite{cheng2022innovae, MOTOHASHI2023122916}) as well as similarity analysis of human ideas (\cite{Just02072024, WANG2023102238}). However, none of these studies propose a system of generative agents itself that can be directly applied to innovation processes performed only by AI. Reference \cite{girotra2023ideas} evaluates the idea generation capabilities of LLMs using zero-shot and few-shot prompting; however, it does not incorporate interactions among multiple agents or agents with internal states. Reference \cite{nisioti2024collective} simulates human cultural evolution by conducting experiments in which groups of LLMs play Little Alchemy 2. However, the primary focus is on knowledge graph exploration rather than technological or corporate innovation. Additionally, it does not involve the design of agents with internal states.

Apart from technological and corporate innovation, numerous studies have examined the creative capabilities of generative AI, particularly in areas such as image generation. For instance, \cite{colombo2023exploring} attempts to analyze the influence of cultural biases on human innovation process using image generation AI. However, it does not utilize LLM agents with internal states. Additionally, \cite{imasato2024} simulates the \textit{systems model of creativity} as dynamic interactions among three entities: visual agent artist that creates paintings, field critic that evaluates them, and domain that indicates evaluation trends. Although \cite{imasato2024} does not focus on technological or corporate innovation, the roles of the external \textit{field critic} are similar to the introspection function of this study.

\subsection{AGENT ARCHITECTURE}

The agent architecture in this study is inspired by several studies on LLM-based agents, including multi-agent systems (for review articles, see \cite{cheng2024survey, guo2024survey, huang2024survey, li2024survey, sun2024survey, xi2025survey, wang2024survey, zhang2024surveymemory, zhao2023survey}). First, many studies have incorporated a memory module (\cite{chuang2024opinion, ge2023llOS, hu2023chatdb, kaiya2023LyfeAgents, li2023metaagents, lin2023agentsims, liu2023thinkinmemory, lu2023memochat, modarressi2024retllm, Omirgaliyev, packer2024memgpt, park2023generativeagents,  qian2024chatdev, ruan2023tptu, shao2023character, Shinn2023, wang2023voyager,williams2023epidemic, yao2024retroformer, zhang2024building, Zhang2024Recommendation, Zhao2024Expel, zheng2024synaps, zhu2023ghost}). For LLM-based agents, a memory system is crucial not only for mimicking human cognitive processes but also for facilitating self-evolution through experience accumulation or knowledge extraction (\cite{zhang2024surveymemory}). In this study, the memory module plays a key role in ensuring context-awareness and consistency in dialogue, as well as promoting the self-evolution of thought. In particular, as in studies such as \cite{chuang2024opinion, li2023metaagents, park2023generativeagents, yao2024retroformer, Zhang2024Recommendation}, the memory module of this study stores reflections in addition to observations (documents and conversations), aiming to enhance idea refinement and knowledge extraction through the recurrent process of thought and dialogue. On the other hand, since the proposed framework is mainly designed for short-term discussions on focused topics, it does not introduce memory operations such as selection based on recency, importance, relevance, or similarity (\cite{hu2023chatdb, kaiya2023LyfeAgents, li2023metaagents, Omirgaliyev, park2023generativeagents, Zhang2024Recommendation}), embedding vectorization (\cite{hu2023chatdb, kaiya2023LyfeAgents, lin2023agentsims, liu2023thinkinmemory, packer2024memgpt, Zhang2024Recommendation, zheng2024synaps}), or distinction between short-term and long-term memory (\cite{kaiya2023LyfeAgents, qian2024chatdev, Shinn2023, yao2024retroformer}).

Apart from memory modules, numerous studies have focused on designing the internal processes of agents that serve as the basis for decision-making (\cite{chuang2024opinion, kaiya2023LyfeAgents, li2023metaagents, Li2024LongShortTermReasoning, Li_2023ToM, lin2023agentsims, Self-Refine, park2023generativeagents, Shinn2023,  yao2023react, zhang2024building, zhang2024cumulativereasoning}). Among these, this study is characterized by three key features. First, the internal state module in this study adopts a recursive structure to update agents' current thoughts iteratively. Due to this structure, even without explicit memory operations such as memory selection, critical insights from past reflections or conversations are refined into higher-level inferences, strongly influencing agents' future reflections and statements. Similar structures are employed in studies such as \cite{kaiya2023LyfeAgents, Li_2023ToM}. Second, this study incorporates an introspection process within the internal state module, functioning as an inner critic. Other studies that have integrated inner critic mechanisms include the self-verification module in \cite{wang2023voyager}, reflection mechanism in \cite{li2023metaagents}, \textit{Refine} step in \cite{Self-Refine}, and evaluator's role in \cite{Shinn2023}. Additionally, some studies have positioned the critic function externally to the agent, such as the \textit{Critic-In-The-Loop} mechanism in \cite{CAMEL} and \textit{Field critic} in \cite{imasato2024}. Third, this study adopts a structure in which diverse intrinsic motivations are reflected in the idea generation process. This ensures that the dialogue process balances both innovation and cooperation, which are essential components of real-world innovation processes. To my knowledge, no existing study has adopted such a structure for idea generation.

\section{CONCLUSIONS}
This study examines whether collective reasoning among generative agents can facilitate novel and coherent thinking that leads to innovation. To explore this, it proposes a new LLM-empowered framework designed to enable reflection and interaction among agents, replicating analogy-driven innovation. It demonstrates, using Dyson's bladeless fan as a case study, that an architecture recursively processing agents' internal states, combined with a dialogue scheme designed for analogy-driven innovation, enables the construction and sharing of coherent and comprehensive concepts.

Future research directions include: (1) examining agents' behavior under more complex organizational structures, (2) utilizing external tools such as CAD and integrating multimodal capabilities into generative agents, and (3) demonstrating collaboration with human teams through partnerships with real-world companies.

% -----------------------------------------
% References
% -----------------------------------------

%\printbibliography[title={REFERENCES}]
\bibliography{GAI_IEEE_OJCS}

\begin{IEEEbiography}[{\includegraphics[width=1in,height=1.25in,clip,keepaspectratio]{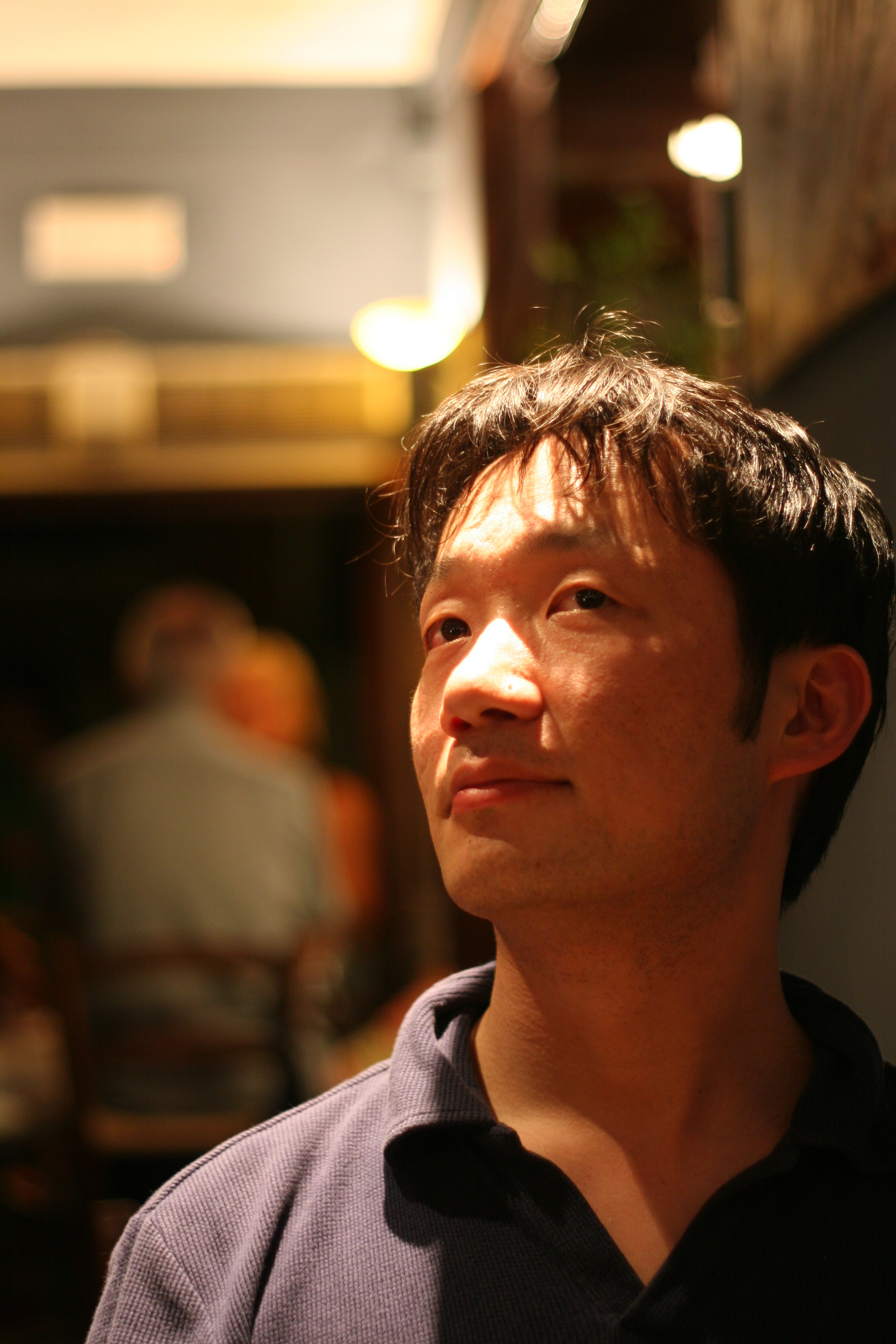}}]
{Masahiro Sato}~(Member, IEEE)~received a Master of Arts and Sciences from the University of Tokyo in 2001, an M.A. in Economics from Georgetown University in 2006, and a Ph.D. in Economics from Kyoto University in 2015. He worked for the Japanese government as a policy officer from 2001 2011, and from 2014 to 2017, and served as an Associate Professor at Kyoto University from 2011 to 2014. He is currently an Associate Professor at the Graduate School of International Cultural Studies, Tohoku University. His research focuses on economic theory and machine learning.
\end{IEEEbiography}

\vfill\pagebreak

\end{document}